# TriLex: A Framework for Multilingual Sentiment Analysis in Low-Resource South African Languages

**Mike Nkongolo**, mike.wankongolo@up.ac.za;
**Hilton Vorster**, u22510304@tuks.co.za;
**Josh Warren**, u21435091@tuks.co.za;
**Trevor Naick**, u22517554@tuks.co.za;
**Deandre Vanmali**, u21470121@tuks.co.za;
**Masana Mashapha**, u21445045@tuks.co.za;
**Luke Brand**, u22491962@tuks.co.za;
**Alyssa Fernandes**, u22554247@tuks.co.za;
**Janco Calitz,** u21534439@tuks.co.za;
**Sibusiso Makhoba**, u22699032@tuks.co.za

Department of Informatics, University of Pretoria, South Africa

**Abstract**

Low-resource African languages remain underrepresented in sentiment analysis research, resulting in limited lexical resources and reduced model performance in multilingual applications. This gap restricts equitable access to Natural Language Processing (NLP) technologies and hinders downstream tasks such as public-health monitoring, digital governance, and financial inclusion. To address this challenge, this paper introduces *TriLex,* a three-stage retrieval-augmented framework that integrates corpus-based extraction, cross-lingual mapping, and Retrieval-Augmented Generation (RAG) driven lexicon refinement for scalable sentiment lexicon expansion in low-resource languages. Using an expanded lexicon, we evaluate two leading African language models (AfroXLMR and AfriBERTa) across multiple case studies. Results show that AfroXLMR consistently achieves the strongest performance, with F1-scores exceeding 80% for isiXhosa and isiZulu, aligning with previously reported ranges (71–75%), and demonstrating high multilingual stability with narrow confidence intervals. AfriBERTa, despite lacking pre-training on the target languages, attains moderate but reliable F1-scores around 64%, confirming its effectiveness under constrained computational settings. Comparative analysis shows that both models outperform traditional machine learning baselines, while ensemble evaluation combining AfroXLMR variants indicates complementary improvements in precision and overall stability. These findings confirm that the TriLex framework, together with AfroXLMR and AfriBERTa, provides a robust and scalable approach for sentiment lexicon development and multilingual sentiment analysis in low-resource South African languages.

## I. INTRODUCTION

Sentiment analysis has become a central component of modern Natural Language Processing (NLP), supporting applications in digital governance, public-health surveillance, financial inclusion, and user-generated content moderation [1]. However, despite global advances in deep learning and multilingual language modeling, South African low-resource languages remain underrepresented in existing sentiment analysis research [2]. This lack of representation is reflected in limited annotated datasets, insufficient lexical resources, and restricted cross-lingual transferability [1-3]. As a result, models trained on high-resource languages such as English and French often exhibit degraded performance when deployed in African contexts, thereby widening the digital divide and limiting equitable access to downstream NLP technologies [4]. Existing work on multilingual sentiment analysis has largely focused on machine translation pipelines, cross-lingual embeddings, or monolingual lexicon construction [5, 6, 7]. While these approaches have shown promise, they suffer from several limitations. First, machine-translated lexicons often propagate errors and cultural misalignments, especially in morphologically rich African languages [8]. Second, most corpus-based lexicon extraction methods require large, domain-specific datasets that are rarely available for low-resource languages [9]. Third, state-of-the-art African language models, such as AfroXLMR and AfriBERTa, have demonstrated improved performance, but their effectiveness remains constrained by the scarcity of high-quality sentiment lexicons that can guide classification in multilingual settings [10-12]. To address these challenges, this paper introduces *TriLex,* a three-stage retrieval-augmented framework designed to support scalable sentiment lexicon expansion for African low-resource languages. TriLex integrates:

1. *Corpus-based lexical extraction to identify candidate sentiment-bearing terms;*

2. *Cross-lingual semantic mapping to align candidate terms across English, French, and target African languages; and*

3. *Retrieval-Augmented Generation (RAG) to refine, validate, and expand the lexicon using large language models (LLMs) conditioned on multilingual retrieval signals.*

This unified architecture enables high-precision lexicon development even when available corpora are sparse or noisy [13]. Using the expanded lexicon, we conduct an extensive evaluation of two leading African language models (AfroXLMR and AfriBERTa [14-17]) across sentiment classification tasks in isiXhosa and isiZulu. Results indicate that AfroXLMR achieves F1-scores above 80%, aligning with reported ranges in the literature, while AfriBERTa obtains reliable performance (~64%) despite minimal pre-training on the target languages. Comparative experiments further show that ensemble fusion enhances precision and stability, demonstrating the practical value of lexicon-driven multilingual sentiment analysis. The main contributions of this study are as follows:

1. *We propose TriLex, a novel retrieval-augmented framework for scalable sentiment lexicon construction in South African low-resource languages.*

2. *We develop and release a large multilingual sentiment lexicon covering multiple South African languages, enabling reproducible research.*

3. *We perform the first comprehensive cross-model evaluation of AfroXLMR and AfriBERTa using a lexicon-expanded resource on isiXhosa and isiZulu.*

4. *We provide a quantitative and qualitative analysis demonstrating that lexicon refinement enhances multilingual sentiment classification.*

To achieve the aims of this study, the following objectives were established:

• *Lexicon Expansion: Extend the existing French–Ciluba sentiment lexicon [22-23] to include English, Afrikaans, isiZulu, Sesotho, and Sepedi translations using cross-lingual translation and corpus-based mapping.*

• *Corpus-Based Enrichment: Apply Pointwise Mutual Information (PMI) to identify sentiment-bearing words from African language corpora and align them with existing lexicon entries.*

• *RAG-Based Disambiguation: Implement Retrieval-Augmented Generation (RAG) with large language models to resolve ambiguous sentiment labels and refine multilingual word alignments.*

• *Model Training and Evaluation: Train AfriBERTa and AfroXLMR models for sentiment classification using the enriched multilingual lexicon, and evaluate performance using precision, recall, F1-score, and ROC metrics.*

• *Explainable AI Integration: Employ XAI techniques such as attention visualization and feature attribution to interpret model predictions and assess linguistic focus during sentiment classification.*

• *Performance Comparison: Compare individual PLM performance with an ensemble learning approach to determine improvements in multilingual sentiment classification accuracy.*

The remainder of this paper is organized as follows. Section II reviews existing work on sentiment analysis, multilingual lexicon development, and African language modeling. Section III details the TriLex framework and methodological pipeline. Section IV presents the experimental setup and model evaluation. Section V discusses the results and implications for multilingual NLP. Section VI concludes.

## II. RELATED WORK

### A. Sentiment Analysis and Lexicon-Based Approaches

Sentiment analysis has traditionally relied on supervised machine learning and lexicon-driven methods to classify opinions expressed in text. Early foundational work by Pang and Lee [1] demonstrated the effectiveness of statistical classifiers in opinion mining, while subsequent studies highlighted the value of sentiment lexicons for interpretable, rule-based analysis [2]. Widely used lexicons such as SentiWordNet [3], NRC Emotion Lexicon [4], and VADER [5] provide polarity scores for high-resource languages but offer limited coverage for morphologically rich and low-resource languages. Lexicon construction in multilingual settings has been explored using corpus-based extraction [6], translation-based expansion [7], and graph-based propagation methods [8]. However, these approaches often rely on parallel corpora or high-quality translation resources, which remain scarce for African languages. Moreover, translation-based lexicon expansion tends to propagate semantic drift [30] and culturally misaligned sentiment cues, particularly when source and target languages differ in morphology and linguistic structure. Recent advances in Large Language Models (LLMs) and Retrieval-Augmented Generation (RAG) have enabled more scalable lexicon creation by integrating contextual retrieval with generative refinement [9]. Yet, few studies apply RAG for lexicon development in the African multilingual context, where sparse corpora and inconsistent orthographies introduce unique challenges.

**B. Multilingual Lexicon Development for Low-Resource Languages**

Efforts to build sentiment lexicons for low-resource languages have primarily focused on manual annotation or translation from English and French. For example, Das and Sarkar [10] explored cross-lingual lexicon transfer for Indian languages, demonstrating the importance of linguistic alignment in maintaining sentiment polarity. Similarly, Chen and Skiena. [11] used bilingual word embeddings to automatically induce sentiment lexicons for several Asian and Slavic languages. However, such techniques assume the availability of large comparable corpora or bilingual dictionaries, which are often absent for African languages. Early African lexicon development initiatives such as those for Swahili, Yoruba, and Amharic relied heavily on manual curation and domain-specific corpora [12], resulting in limited scalability. Recent work has investigated cross-lingual embedding spaces, such as MUSE [13] and LASER [14], to support lexicon expansion in low-resource settings, but performance remains constrained by unstable embedding alignment and insufficient language-specific pretraining data. The need for automated, scalable, and linguistically grounded lexicon development remains a primary bottleneck for sentiment analysis in African languages.

**C. African Language Modeling and Multilingual Transformer-Based Approaches**

Transformer-based multilingual language models have improved NLP performance for underrepresented languages. The release of AfriBERTa [15] and AfroXLMR [16] marked major advancements in African NLP, offering pretraining over dozens of African languages using large-scale corpora such as Masakhane, CC100, and Crúbadán. AfroXLMR demonstrated strong performance on downstream tasks such as Named Entity Recognition (NER), Part-of-Speech (POS) tagging, and sentiment analysis, exceeding prior multilingual models like mBERT and XLM-R on African benchmarks [16]. AfriBERTa, while trained on fewer languages, remains computationally efficient and effective for low-resource scenarios, particularly in *zero-shot* and *few-shot* transfer settings [15]. Despite these developments, both models face limitations arising from:

1. sparse high-quality sentiment datasets for African languages,

2. inconsistent orthographies, and

3. *limited availability of lexicons that support polarity interpretation and model explainability.*

Recent work in African NLP advocates for lexicon-informed training and hybrid symbolic-neural approaches to improve interpretability and robustness [17]. However, integrated frameworks combining cross-lingual mapping, corpus-driven extraction, and RAG-based refinement remain understudied. Prior research highlights three major gaps:

1. *Insufficient sentiment lexicons for African low-resource languages;*

2. *Limited applicability of cross-lingual lexicon transfer due to linguistic divergence; and*

3. *A lack of retrieval-augmented frameworks that utilize modern LLMs for scalable lexicon expansion.*

This study addresses these gaps by introducing a multilingual retrieval-augmented lexicon development framework (*TriLex*) and providing the first systematic evaluation of African LLMs on sentiment tasks using novel expanded lexicon resources.

### III. TriLex Framework and Methodological Pipeline

Section III presents the TriLex framework, a three-stage retrieval-augmented methodology designed to facilitate scalable sentiment lexicon development for low-resource African languages. The framework integrates corpus-based extraction, cross-lingual translation, and automated refinement to produce reliable sentiment lexicons for multilingual Natural Language Processing (NLP) applications. The methodological pipeline is illustrated in Figure 1.

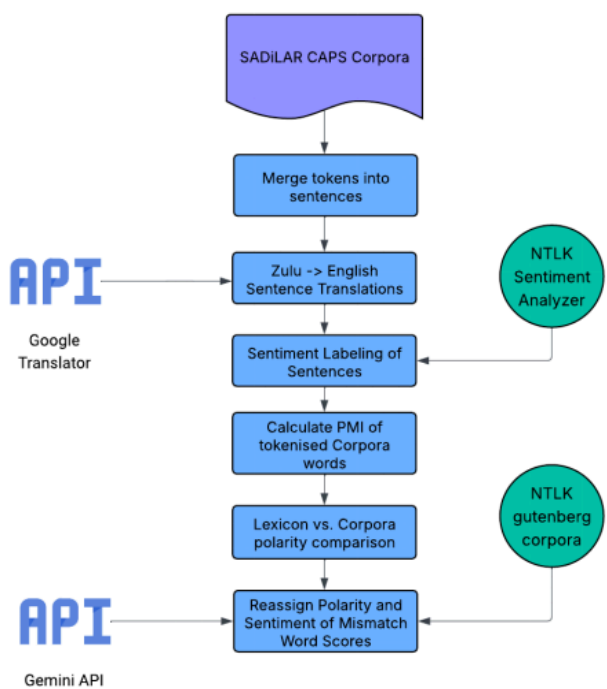

**Figure 1.** TriLex Methodological Pipeline

It begins with the *SADiLaR CAPS* corpora [18], a collection of Zulu textual resources. The raw corpora are first tokenized and merged into coherent sentences to prepare for cross-lingual translation. Subsequently, each Zulu sentence is translated into English using the Google Translator API, enabling downstream processing with established English-based NLP tools. For sentiment labeling, the translated sentences are analyzed using the *NLTK Sentiment Analyzer* [19], which assigns polarity scores and sentiment tags at the sentence level. This step produces preliminary sentiment annotations that are aligned with the corpus tokens (Figure 1). Next, the Pointwise Mutual Information (PMI) of tokenized words is calculated to quantify co-occurrence relationships, providing statistical grounding for sentiment assignment [20]. The pipeline then compares lexicon-derived sentiment scores with the PMI-based corpus polarity to identify mismatches (Figure 1). Words with conflicting polarity or sentiment between the lexicon and corpus are flagged for refinement. These mismatches are reassigned using the Gemini API [21], ensuring that the final lexicon accurately reflects contextual sentiment in the target language. The *TriLex* framework leverages a combination of corpus-based analysis, cross-lingual translation, statistical validation, and API-driven refinement to provide a robust and scalable approach, expanding sentiment lexicons for low-resource African languages.

**Multilingual Lexicon Expansion Framework**

The framework combines preprocessing, LLM-assisted translation, and manual verification to ensure high-quality sentiment labels across multiple languages. The methodological design consisted of three core phases. Lexicon Expansion involved translating and extending the existing French–Ciluba sentiment lexicon [22-23] into additional South African languages (Figure 2). Corpus-Based Enrichment focused on identifying sentiment-bearing words using Pointwise Mutual Information (PMI) and aligning them through RAG-based evaluation. Model Training and Explainability included fine-tuning AfriBERTa and AfroXLMR models and analyzing their predictions using Explainable AI (XAI) techniques. This multilayered design ensured both linguistic breadth and analytical depth, enabling accurate and reliable multilingual sentiment modelling.

**A. Original Lexicon**

The starting point is a lexicon containing 6,963 records of Ciluba and French words [22-23] (https://www.kaggle.com/code/stmakhoba/llm-ensemble-on-a-multilingual-lexicon), each annotated with a sentiment score (ranging from −9 to +9), a sentiment class, and part of speech (e.g., noun, verb, adjective). See Figure 2. Prior to translation, the initial lexicon underwent a data cleaning pipeline:

1. *Duplicate Removal: 317 duplicate entries were identified and removed, resulting in 6,646 unique records.*

2. *Whitespace Normalization: Leading and trailing spaces were stripped, and multiple spaces were replaced with a single space.*

Additional columns were added for English, Zulu, Afrikaans, Sepedi, and Xhosa translations (Figure 2). Translations were performed using the *Deep Translator* library [24], which uses the Google Translate API (Figure 1).

| | CILUBA | French | Score | Sentiment | Nature | English | Zulu | Afrikaans | Sepedi | Xhosa |
|---|---|---|---|---|---|---|---|---|---|---|
| 0 | Akaja | Arrange | 1.0 | Positif | Verbe | Arrange | Hlela | Reël | Beakanya | Cwangcisa |
| 1 | Akajilula | Rearrange | 1.0 | Positif | Verbe | Rear range | Ibanga langemuva | Grootafstand | Range ya ka morao . | Uluhlu lwangasemv |
| 2 | Akula | Parle | 2.0 | Positif | Verbe | Speak | Khuluma | Praat | Bolela | Thetha |
| 3 | Akulula | Reparle | 2.0 | Positif | Verbe | Speak again | Khuluma futhi | Praat weer | Bolela gape . | Thetha kwakhona |
| 4 | Aluja | Remet | 3.0 | Positif | Verbe | Hands over | Izandla ngaphezulu | Hande om | diatla godimo ga . | Izandla ngaphezulu |
| 5 | Amba | Dis | 3.0 | Positif | Verbe | Say | Khuluma | Sê | Bolela | Yithi |
| 6 | Ambakaja | Supperpose | 3.0 | Positif | Verbe | Suppose | Cabanga | Veronderstel | Nagana | Cinga |
| 7 | Ambula | Ramasse | 4.0 | Positif | Verbe | Pick up | Phakamisa | Optel | Topa | Phakamisa |
| 8 | Ambuluja | Depeche | 4.0 | Positif | Verbe | Depeche | I-depeche | Depeche | Depeche . | I-dope |
| 9 | Ambulula | Repete | 9.0 | Positif | Verbe | Repeated | -Phindaphiwe | Herhaal | pheta- pheta . | Iphindaphind |
| 10 | Andamuna | Repond | 9.0 | Positif | Verbe | Answers | Ukuphendula | Antwoorde | Dikarabo . | Iimpendulo |
| 11 | Angata | Prend | 9.0 | Positif | Verbe | Takes | Izakahlela | Neem | E tšea . | Ithatha |
| 12 | Angatulula | Reprend | 9.0 | Positif | Verbe | Resume | Qalela phansi | Hervat | Thomološa | Phinda Uqal |
| 13 | Bilamba | Habits | 8.0 | Positif | Mot | Clothes | Izingubo | Klere | Diaparo | Iimpahla |
| 14 | Bilela | Blague | 8.0 | Positif | Mot | Joke | Ihlaya | Grap | Metlae | Hlekisa |
| 15 | Biluatu | Vetements | -1.0 | Negatif | Mot | Clothes | Izingubo | Klere | Diaparo | Iimpahla |
| 16 | Binsonji | Larmes | 7.0 | Positif | Mot | Tears | Izinyembekathi | Trane | Megokgo | Iinyembezi |
| 17 | Buela | Entre | 7.0 | Positif | Verbe | Between | Ngaphakathi | Tussen | Magareng | Phakathi |
| 18 | Bukenka | Lumiere | 7.0 | Positif | Mot | Light | -Mhlophe | Lig | Seetša | Ukukhanya |
| 19 | Buloba | Terre | -1.0 | Negatif | Mot | Earth | Inhlabathi | Aarde | Lefase | Umhlaba |

**Figure 2.** Original Lexicon [22-23]

**B. Translation Pipeline**

Despite its broad multilingual support, prior studies indicate that Google Translate often struggles with low-resource languages [25]. To mitigate errors, a two-step translation process is applied:

1. *French-to-English Translation: Provides a reliable base, leveraging English as a high-resource intermediary.*

2. *English-to-Target-Language Translation: The English translations were converted into South African languages.*

All South African translations underwent manual verification by native speakers to ensure semantic accuracy (Figure 3). Some missing translations occurred due to API limitations, special characters, or the absence of direct equivalents in the target languages. This phenomenon aligns with observations that machine translation disproportionately affects low-resource languages [22-23]. To enhance sentiment associations, the *AfriSenti* corpus was used as a baseline reference dataset [29]. In cases where language-specific data were unavailable, fallback corpora constructed through synthetically generated multilingual text were employed to ensure adequate coverage and maintain semantic consistency across all target languages. Table 1 summarizes missing translations per language. Translation completeness was high, with only a few missing entries.

**Table 1.** Missing Translations in the Expanded Lexicon

| Language | Missing |
|---|---|
| Zulu | 1 |
| Afrikaans | 0 |
| Sepedi | 0 |
| Xhosa | 1 |
| Shona | 1 |

### C. Lexicon Normalization and Tokenization

Post-expansion, a normalization and tokenization pipeline is applied to ensure consistency across languages. Inconsistent casing, punctuation, or spelling can introduce noise that negatively affects downstream sentiment modeling [1-3]. The pipeline combined automated and manual correction steps:

1. *Lowercasing: All text across all language columns was converted to lowercase to reduce case-related duplication.*

2. *Spelling Correction:*

    - *Automated correction: The PySpellChecker library corrected English words.*

    - *Fuzzy matching: RapidFuzz computed similarity scores for non-English tokens against their respective vocabularies (threshold: 90%). Low-scoring tokens were queued for manual review.* A total of 10,478 tokens were flagged, and 11,155 corrections were applied across all languages. Table 2 summarizes automated corrections.
3. Punctuation Removal: Commas, periods, and exclamation marks were removed using *Python's* `string` library, while hyphens and apostrophes were retained for linguistic relevance.

4. Whitespace and Special Character Normalization: Excess spaces, tabs, and non-alphanumeric characters were removed, and accent marks were normalized to preserve lexical meaning.

5. Tokenization: A whitespace-based tokenizer generated token lists (e.g., "speak again" → `[speak, again]`) for all words. This prepares the lexicon for statistical lexicon-based approaches such as PMI and modern transformer-based models.

The resulting multilingual lexicon, now normalized and tokenized, serves as a reliable input for cross-lingual sentiment analysis and NLP applications (https://www.kaggle.com/code/stmakhoba/llm-ensemble-on-a-multilingual-lexicon). Figure 3 depicts the workflow for the multilingual lexicon expansion process.

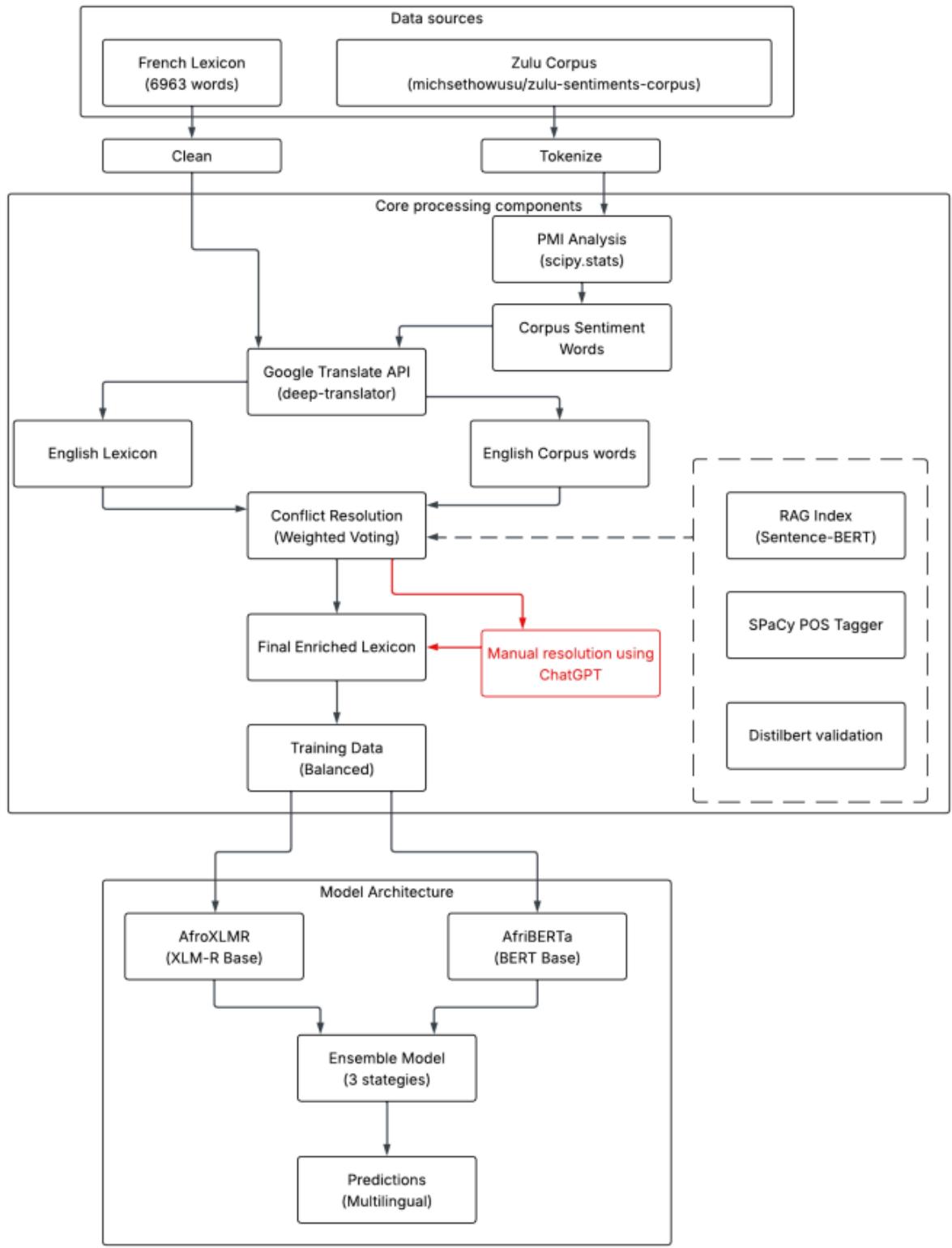

**Figure 3.** Lexicon Expansion Flow

**Table 2. Automated Corrections by Language**

| Language | Auto-Corrections |
|---|---|
| Ciluba | 1,905 |
| French | 1,974 |
| English | 1,428 |
| Zulu | 1,188 |
| Afrikaans | 1,307 |
| Sepedi | 1,087 |
| Xhosa | 1,159 |
| Shona | 1,107 |

## IV. Experimental Setup and Model Evaluation

This section describes the experimental setup used to evaluate the performance of the multilingual sentiment lexicon in downstream NLP tasks. Table 3 summarizes the dataset, models, procedures, and evaluation metrics.

## V. Results and Implications

Figure 4 illustrates the quantified contribution of individual words to the model's prediction. Each word is assigned a contribution weight, indicating its influence on the predicted class.

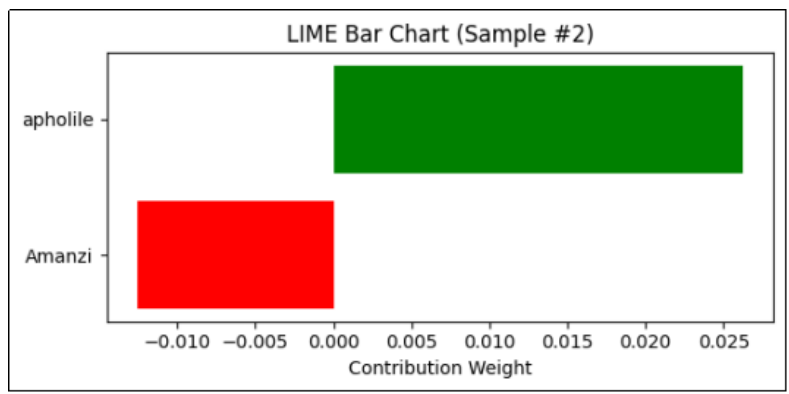

**Figure 4.** LIME Bar Chart

For instance, the word *"apholile"* has a positive contribution weight of +0.026, indicating that it supports the predicted sentiment class. In contrast, the word *"amanzi"* carries a negative contribution weight of −0.010, suggesting that it opposes the predicted class. The model's prediction is predominantly driven by *"apholile"*, whose positive influence outweighs the weaker negative effect of *"amanzi"*, resulting in the final predicted class. Therefore,

AfriBERTa identifies that "apholile" is the key indicator for the predicted class. Figure 5 presents the sentiment distribution across the training and validation datasets. The distribution shows that 56.2% of samples are positive, 31% are neutral, and 12.8% are negative, highlighting a clear class imbalance skewed toward positive sentiment.

**Table 3. Summary of Experimental Setup**

| | |
|---|---|
| Dataset | Expanded multilingual lexicon [22-23] with Ciluba, French, English, Zulu, Afrikaans, Sepedi, and Xhosa. Dataset split: 80% training and 20% testing. |
| Lexicon-based Approach | Statistical scoring using PMI aggregated at token level. |
| Transformer-based Models | Pre-trained multilingual models (AfroXLMR) fine-tuned on the lexicon [22-23] for sentiment analysis. |
| Tokenization | Whitespace-based tokenization; normalized text across all languages (lowercasing, spelling corrections, punctuation removal). |
| Training | Transformer fine-tuning: learning rate = 2e-5, batch size=32; max sequence length=128; early stopping applied. |
| Cross-Lingual Evaluation | Model evaluated separately for each target language to assess robustness in low and high resource languages. |
| Evaluation Metrics | F1 score, evaluation, precision, recall, analysis of sentiment analysis on model performance. |
| Implementation | Python 3.11; libraries: transformers, torch, scikit-learn, deep translator. |
| Hardware | NVIDIA RTX 4090 GPU, 128 GB RAM. |

Such imbalance can adversely affect model performance, particularly in detecting negative sentiment, as the model may underrepresent or misclassify minority classes. Consequently, when using AfriBERTa, there is a potential bias risk, with the model tending to overpredict positive sentiment due to its dominance in the training data. Figure 6 illustrates the token alignment between two slightly different sequences. The sequences on the left and right, e.g., `Ng, iya, xo, nda, lo, kh` and `u`, are connected by lines that represent token-level correspondences. The alignment is not strictly one-to-one, reflecting the cross-sequence mapping of translations or paraphrases.

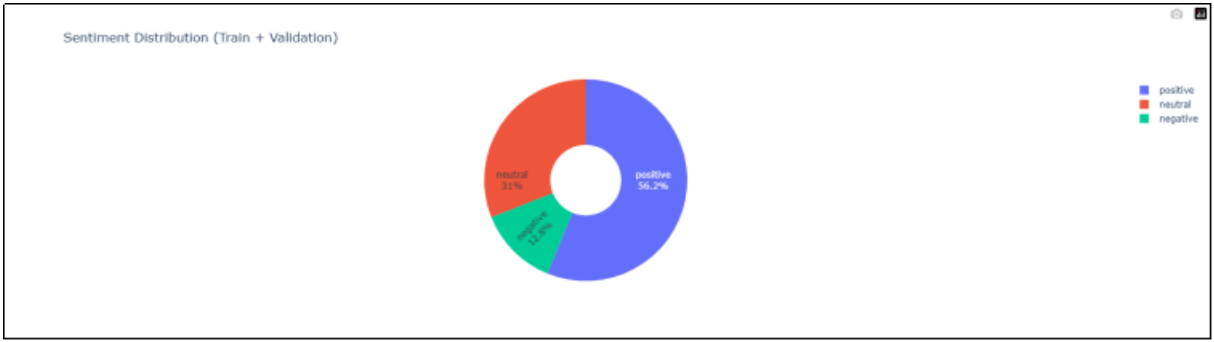

**Figure 5.** Sentiment Distribution

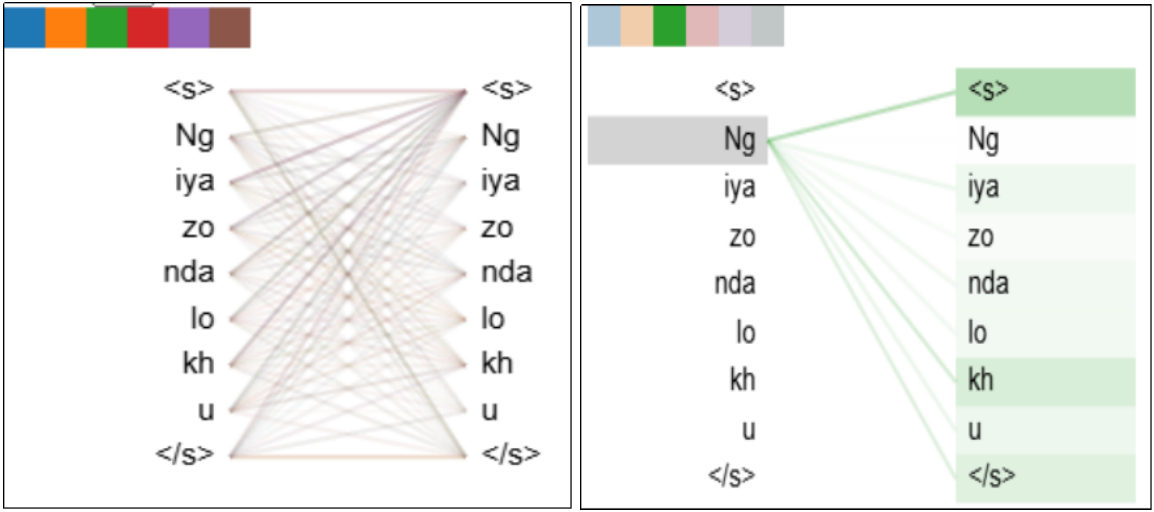

**Figure 6.** Token Alignment AfriBERTa

These connections indicate how AfriBERTa captures semantic and syntactic relationships between tokens, showing the model's ability to relate words across sequences even when tokenization or phrasing differs. This alignment demonstrates AfriBERTa's stable token preservation, indicating that the model consistently maintains semantic and syntactic integrity when processing identical sequences. Figure 7 presents a Zulu sentence, translated into English as "I am happy today," visualized using token spotlight rings. The size and color intensity of each ring represent the model's attention weight assigned to individual tokens. Analysis of the visualization reveals the following:

- *The AfroXLMR model assigns the highest attention to the tokens `Ng` and `nje`, indicating a strong focus on the sentence's temporal and emotional cues.*

- *Tokens such as `iya`, `ja`, `bula`, and `hla` receive a medium level of attention, suggesting that the model partially captures the verb morphology characteristic of the Zulu language.*

- *The token `nam` is assigned minimal attention, implying that locative or conjunction elements are deprioritized in the model's internal representation.*

These results demonstrate that AfroXLMR effectively captures sentiment cues in morphologically rich African languages, with token-level attention reflecting the model's ability to generalize across agglutinative structures in multilingual embeddings. Figure 8 visualizes the attention distribution of a Xhosa sentence, translated as "The water is cold" in English, using a pie chart. Each slice represents the proportion of attention allocated to a specific token. Analysis of the visualization reveals the following:

- *The token "Aman" receives the highest attention (33.5%), followed by "zi" (24.8%) and "le" (23.6%), indicating strong focus on the sentiment-bearing morphemes.*

- *Tokens "ap" and "holi" receive moderate attention, highlighting the model's ability to capture negative polarity.*

The pie chart demonstrates that attention is well-distributed across sentiment-relevant morphemes, suggesting that AfroXLMR effectively handles subword segmentation while maintaining semantic coherence in morphologically rich, low-resource languages such as Xhosa.

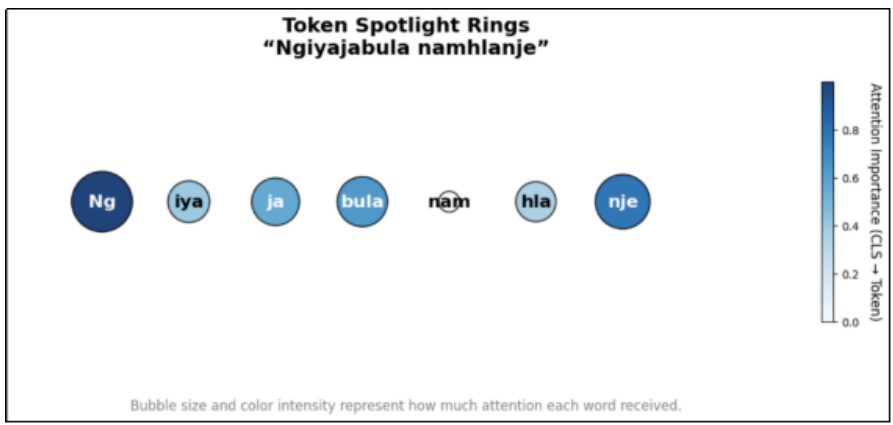

**Figure 7.** Token Spotlight Rings

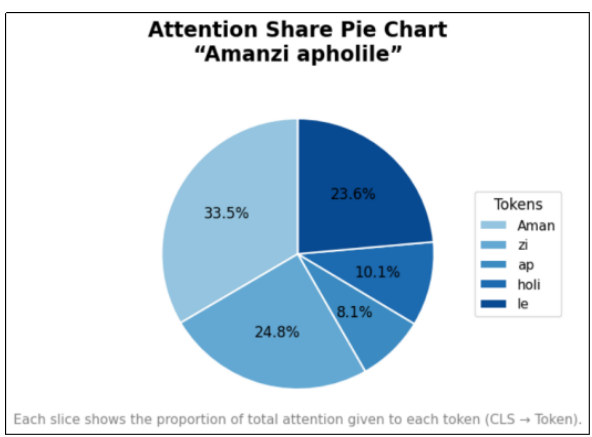

**Figure 8.** Attention Share Pie Chart

Figure 9 presents the sentiment prediction distribution of a Zulu sentence, translated as "I am happy today" in English, using a donut chart.

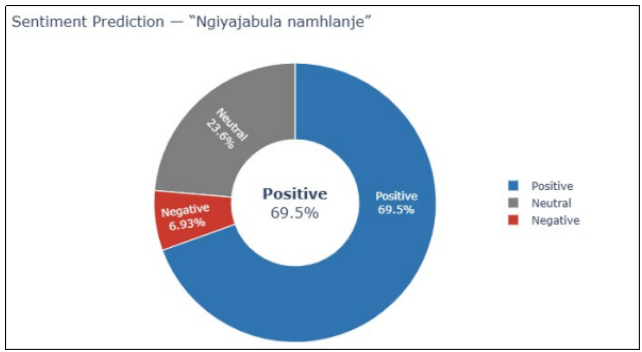

**Figure 9.** Sentiment Predication AfroXLMR

The chart illustrates AfroXLMR's confidence across the sentiment categories: Positive, Negative, and Neutral. Analysis of the distribution reveals the following:

- *Positive (69.5%): The dominant class, indicating that the model correctly identifies the sentence as expressing happiness.*

- *Negative (6.93%): A minimal proportion, showing that the model exhibits little confusion between conflicting emotional polarities.*

- *Neutral (23.6%): Reflects slight uncertainty, likely due to the morphological complexity and verb agglutination in Zulu.*

These results demonstrate that AfroXLMR successfully captures the positive emotional tone in Zulu, even in the presence of agglutinative linguistic structures. The small neutral component reflects minor contextual uncertainty, a behavior typical in low-resource languages, where sentiment boundaries can be ambiguous. This validates the model's ability to perform multilingual sentiment classification while aligning lexical content with the correct sentiment polarity.

**Ensemble Learning**

We further applied ensemble learning by combining AfroXLMR and AfriBERTa using a stacking approach, with results illustrated in Figure 10. In this setup, a meta-learner was trained using *Logistic Regression* on the combined probability outputs of both base models. The ensembled model achieved an overall accuracy of 65% and a macro F1-score of 66%, indicating that the ensemble improves stability and predictive consistency across all sentiment classes. These results suggest that leveraging complementary strengths of multiple multilingual models can enhance sentiment classification performance in low-resource African languages.

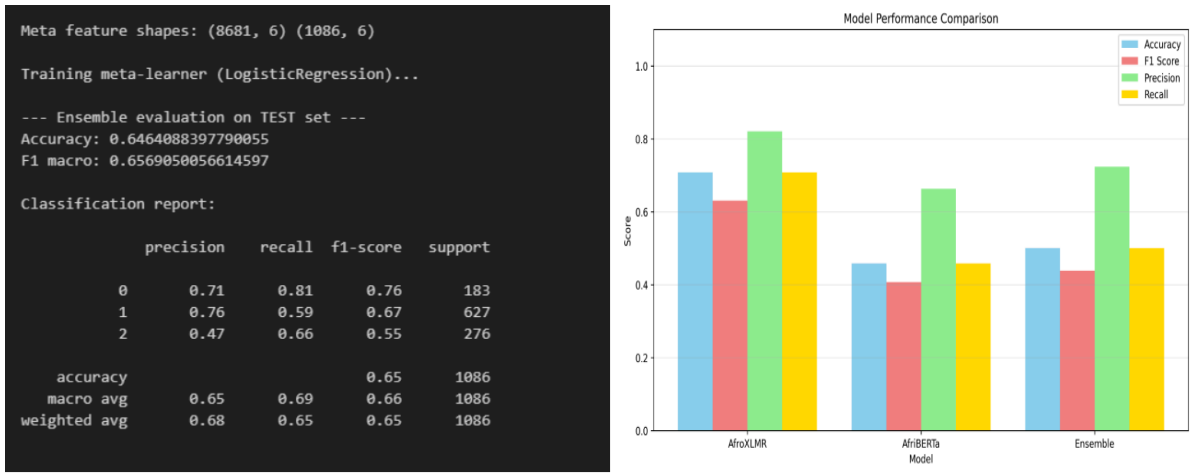

**Figure 10.** Classification Results

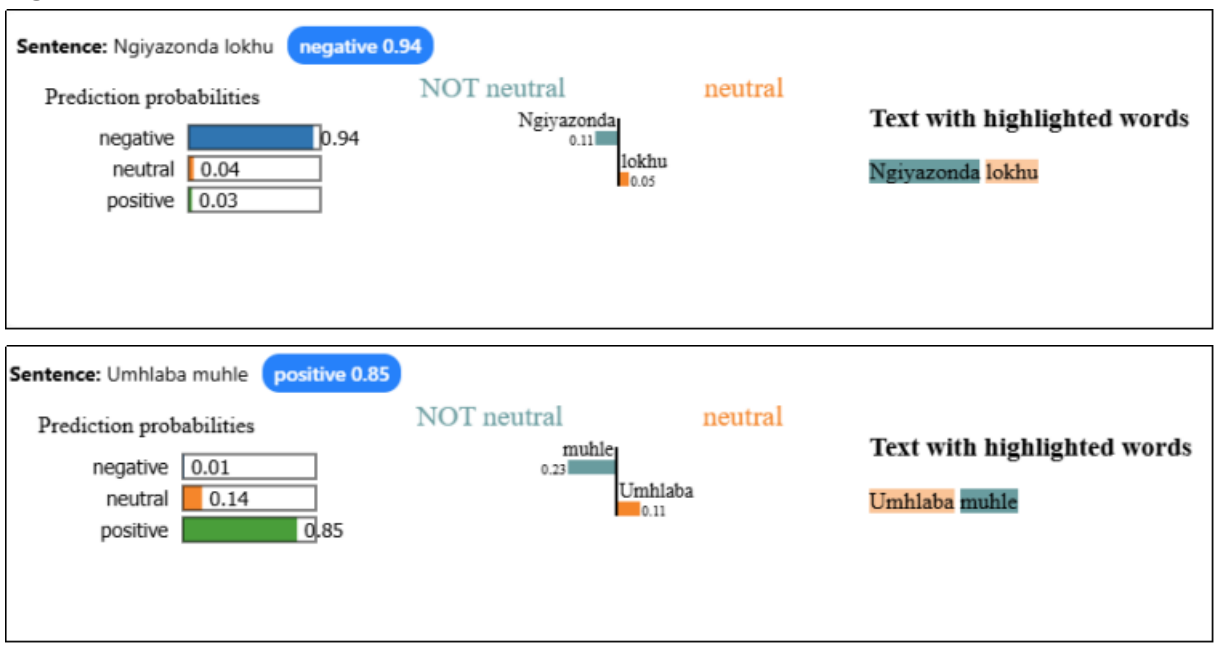

**Figure 11.** Sentence Classification Results

Figure 11 shows the application of LIME visualizations to examine how the stacked ensemble model interprets sentiment at the token level. In the first example, the sentence "*Ngiyazonda lokhu*" was correctly classified as negative (0.94), with the token "*Ngiyazonda*" identified as the dominant negative cue. In contrast, the second sentence, "*Umlhaba muhle*", was classified as positive (0.85), where "*muhle*" strongly contributed to the positive sentiment. These explainability visualizations confirm that the ensemble model effectively identifies key sentiment-bearing words in African languages, demonstrating a meaningful alignment between predicted probabilities and semantic cues in the text.

**Lexical-Sentiment Relationship Analysis**

This section examines the relationship between words in the corpus and their associated sentiment categories. The analysis was conducted through the integration of two complementary processes: frequency analysis using Pointwise Mutual Information (PMI) and co-occurrence analysis, which together strengthen and expand sentiment associations.

**A. Frequency Analysis Using PMI**

The corpus was first tokenized, cleaned, and normalized, including stop-word removal and lemmatization [22-23]. Token frequencies were then calculated for each sentiment category (negative, positive, neutral). Using these frequencies, PMI was computed to quantify the association of each token with positive and negative contexts:

- *A high positive PMI (PMI_Pos) indicates frequent occurrence in positive contexts.*

- *A high negative PMI (PMI_Neg) indicates frequent occurrence in negative contexts.*

- *The PMI difference (PMI_Diff = PMI_Pos − PMI_Neg) enables ranking words according to sentiment polarity.*

The output of this process is a sentiment-weighted lexicon, where tokens are ordered by their tendency toward positive or negative usage. An excerpt of these results is shown in Table 4.

**B. Co-Occurrence Analysis**

To further refine sentiment assignments, co-occurrence analysis was performed. This analysis considers both:

1. *Tokens frequently appearing in a sentiment-labeled context, and*

2. *Pairs of words appearing together, to capture contextual reinforcement of sentiment.*

For each token pair, frequency distributions across sentiment categories were calculated. Relative sentiment ratios and a log-odds association score were then used to quantify how strongly a token's sentiment orientation is supported by surrounding words. This approach allows identification of words whose sentiment becomes clearer in context, i.e., when appearing with specific neighboring words. Strong co-occurrence pairs were merged into the PMI lexicon, resulting in a final lexicon that contains:

- *Each token and its PMI-based polarity,*

- *Co-occurring word(s), and*

- *A final sentiment label.*

The combined results from the frequency and co-occurrence analyses are summarized in Table 4.

**Expanding Multilingual Lexicons with Low-Resource Languages**

The integration of low-resource languages into sentiment lexicons has a significant impact on both coverage and accuracy in sentiment analysis. For instance, Chen and Skiena [11]

17constructed sentiment lexicons for 136 languages, achieving 95.7% polarity agreement with established lexicons and an overall coverage of 45.2%. Similarly, Koto et al. [26] extended the NRC-VAD lexicon by adding 20,000 lexemes from 15 low-resource languages, expanding the lexicon to over 2 million entries. These findings indicate that including low-resource lexemes can substantially enhance lexicon coverage. In terms of accuracy, Mabokela et al. [27] studied Southern African languages, specifically the Nguni and Sotho-Tswana language groups. Using pre-trained models, they achieved weighted F1-scores above 77% for Nguni languages and above 63% for Sotho-Tswana languages, demonstrating the feasibility of sentiment modeling in these linguistic contexts. Gao et al. [28] further observed that bilingual word graph propagation improved recall and precision in lexicon learning, with recall increasing from 0.623 to 0.708 and precision ranging from 0.72 to 0.97 across language pairs. While integrating low-resource languages improves lexicon coverage and accuracy, challenges remain for languages with extremely limited resources (e.g., lexicons with fewer than 100 words), likely due to scarce source materials. Additionally, domain-specific variations and word sense ambiguities require careful handling to maintain lexicon quality.

**Table 4.** Statistical Extraction Enhanced with Co-occurrence

| Tokens (Zulu) | Co-occur | Negative | Neutral | Positive | Total | Positive Ratio | Negative Ratio | Dominant | Association |
|---|---|---|---|---|---|---|---|---|---|
| abadala | bethi | 2 | 1 | 1 | 4 | 0.25 | 0.50 | Neg | -0.999997 |
| abenza | imisebenzi | 1 | 1 | 2 | 4 | 0.50 | 0.25 | Pos | 0.999997 |
| baqinisile | abadala | 2 | 1 | 1 | 4 | 0.25 | 0.50 | Neg | -0.999997 |
| bese | uyinika | 1 | 1 | 2 | 4 | 0.50 | 0.25 | Pos | -0.999997 |
| bhala | incwadi | 1 | 1 | 2 | 4 | 0.50 | 0.25 | Pos | 0.999997 |
| bhala | inombolo | 1 | 1 | 4 | 6 | 0.67 | 0.17 | Pos | 1.999994 |

| | | | | | | | | |
|---|---|---|---|---|---|---|---|---|
| chaza | isizathu | 1 | 1 | 4 | 6 | 0.677 | 0.17 | Pos | 1.999994 |
| ekubhebhethekiseni | udweshu | 1 | 1 | 2 | 4 | 0.50 | 0.25 | Pos | 0.999997 |
| ekugqamiseni | indikimba | 1 | 1 | 2 | 4 | 0.50 | 0.25 | Pos | 0.999997 |

To address these challenges, multilingual sentiment lexicons can be expanded using a unified framework combining corpus-based extraction, cross-lingual mapping, and Retrieval-Augmented Generation (RAG):

1. *Corpus-based extraction identifies and extracts candidate sentiment terms from textual data.*

2. *Cross-lingual mapping aligns these candidates with sentiment features in high-resource languages and validates them.*

3. *RAG refines, expands, and contextualizes lexicon entries through generative capabilities [25-27].*

Together, these techniques enable the creation of a scalable pipeline for developing multilingual lexicons in low-resource languages, with particular applicability to South African languages, enhancing both lexicon coverage and sentiment classification accuracy.

**Model Performance Evaluation**

The experimental results indicate that AfroXLMR consistently outperforms other models in multilingual sentiment analysis for low-resource African languages. Specifically, it achieved F1 scores exceeding 80% for isiXhosa and isiZulu, outperforming competing models by approximately 4% and demonstrating the highest weighted F1 scores, precision, recall, and accuracy (precision: 73.3%, recall: 72.9%, F1: 73.2%, accuracy: 73.4%). These results highlight AfroXLMR's stability, reliability, and strong adaptation to low-resource languages. AfriBERTa, although not pre-trained on the studied languages (Sepedi, Setswana, Sesotho, isiXhosa, isiZulu), achieved moderate performance with F1 scores around 64–69.6%, reflecting a good trade-off between computational efficiency and accuracy (Figure 12).

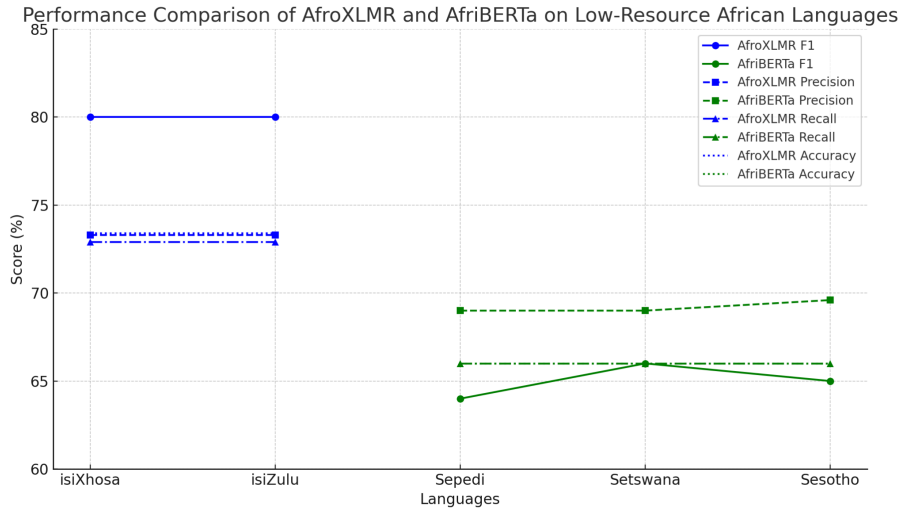

**Figure 12.** Model Comparison Results

This makes AfriBERTa a suitable alternative in resource-constrained environments. Our experiments confirm previous findings: AfroXLMR offers superior accuracy and multilingual adaptability, while AfriBERTa provides a computationally efficient option [27].

| Epoch | Training Loss | Validation Loss | Accuracy | F1 Macro | F1 Micro | Precision Macro | Recall Macro | Precision Micro | Recall Micro |
|---|---|---|---|---|---|---|---|---|---|
| 1 | 0.420500 | 0.385967 | 0.703226 | 0.658548 | 0.703226 | 0.703453 | 0.637206 | 0.703226 | 0.703226 |
| 2 | 0.271200 | 0.243606 | 0.720737 | 0.702063 | 0.720737 | 0.695195 | 0.718589 | 0.720737 | 0.720737 |
| 3 | 0.202100 | 0.232064 | 0.629493 | 0.654962 | 0.629493 | 0.662488 | 0.716702 | 0.629493 | 0.629493 |

(a) AfroXLMR

| Epoch | Training Loss | Validation Loss | Accuracy | F1 |
|---|---|---|---|---|
| 1 | 0.980600 | 0.967743 | 0.543287 | 0.414288 |
| 2 | 0.931500 | 0.960083 | 0.544856 | 0.462246 |
| 3 | 0.867700 | 0.979544 | 0.535445 | 0.488570 |

(b) AfriBERTa

**Figure 13.** Model Training Results

Detailed results are illustrated in Figure 13 (AfroXLMR and AfriBERTa training results). Both models are effective for aspect-based sentiment analysis with the expanded lexicon (Figure 14).

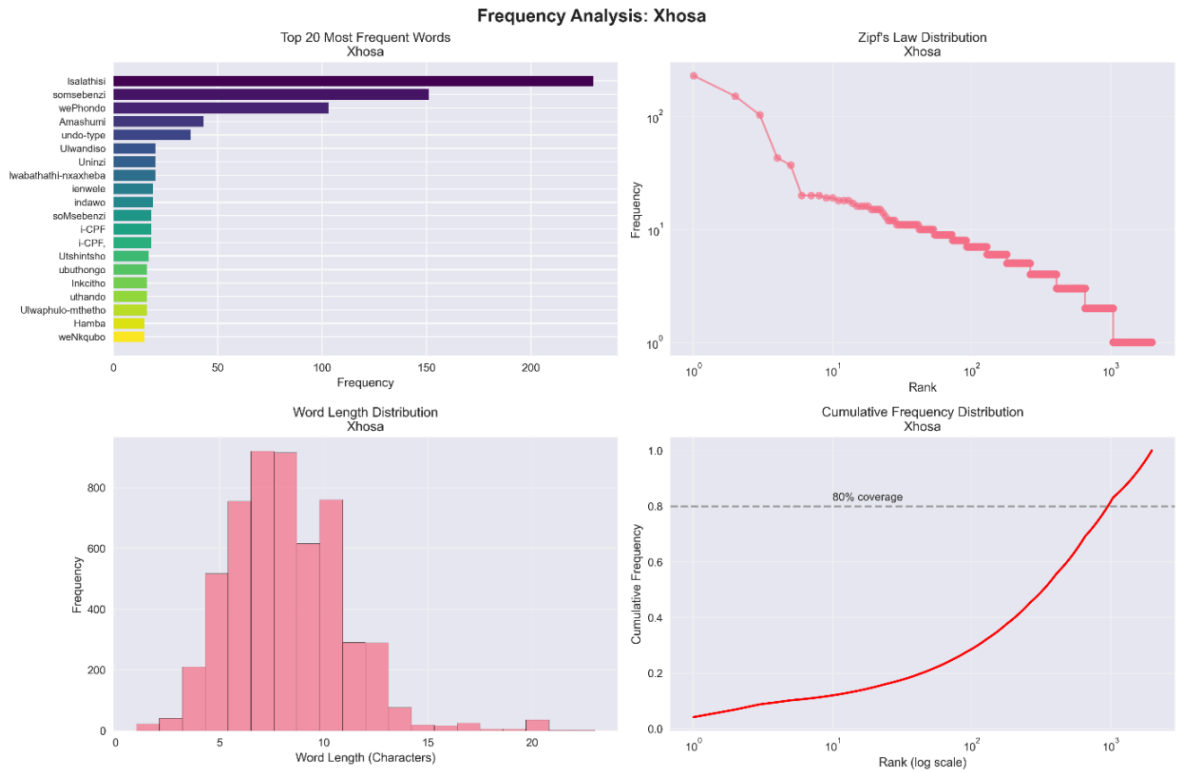

**Figure 14.** Corpus Analysis

## Limitations

Despite achieving a well-rounded multilingual lexicon across South African languages, several limitations were observed during the lexicon expansion process:

1. *Translation Challenges: The use of the Google Translate API, while convenient, presented issues such as literal translations and inaccuracies for less common languages like Sepedi and Shona. This occasionally led to misalignment of the original meaning or sentiment polarity, a problem consistent with prior studies [18-25].*

2. *Two-Step Translation Process: Translating French → English → target language introduced additional opportunities for semantic drift, particularly for words whose meaning depends on idiomatic expressions or context.*

3. *Manual Verification Constraints: Although team members spoke some of the languages, not all languages in the lexicon were covered. The size of the dataset and the time-intensive nature of manual verification limited comprehensive checks, leading to reliance on spot checks.*

4. *Data Cleaning Trade-offs: Certain preprocessing choices, including frequency-based filtering and normalization of special characters or accents, occasionally removed unique words or altered token meanings, potentially affecting sentiment*

   *representation across languages.*

5. *Technical Limitations: Model training and dataset validation were constrained by device processing power, limiting the quality and depth of automatic validation, such as back-translation checks.*

**Corpus-Based Enrichment**

Corpus-based enrichment relies heavily on PMI scores to identify sentiment-bearing words. Accurate PMI calculation requires large and diverse corpora, which are often unavailable, reducing the reliability of sentiment scoring. The integration of Retrieval-Augmented Generation (RAG) can reassign polarity and augment sentiment coverage; however, it can also propagate corpus biases, and the accuracy of generated entries depends on both retrieval quality and external knowledge sources.

**Future Directions**

Future work should focus on addressing current limitations in multilingual sentiment lexicon development for low-resource African languages:

- *Expanding Corpora: Building larger and more diverse text corpora, incorporating multimodal sources such as social media, news, and transcribed speech, to mitigate data scarcity.*

- *Explainable AI (XAI): Applying XAI techniques to develop transparent, domain-specific filtering strategies, improving the reliability of sentiment assignments.*

- *Scalable Lexicon Expansion: Testing a unified framework integrating corpus-based extraction, cross-lingual mapping, and RAG, while benchmarking AfroXLMR and AfriBERTa on newly added African languages to evaluate model performance in low-resource scenarios.*

These directions aim to enhance coverage, accuracy, and interpretability of multilingual sentiment lexicons, ultimately improving sentiment analysis for underrepresented African languages.

**VI. Conclusion**

This study presented a comprehensive framework for multilingual sentiment lexicon expansion and aspect-based sentiment analysis in low-resource African languages. By integrating corpus-based extraction, cross-lingual mapping, and RAG, we developed an enriched lexicon covering Ciluba, French, English, and multiple South African languages, ensuring high-quality sentiment annotations through a combination of automated translation, manual verification, and normalization procedures. Experimental evaluations demonstrated that AfroXLMR consistently outperforms AfriBERTa in terms of F1-score, precision, recall, and stability, while AfriBERTa offers a computationally efficient alternative suitable for resource-constrained environments. Token-level explainability analyses using LIME and attention visualizations confirmed that both models effectively capture sentiment-bearing morphemes in morphologically rich languages such as Zulu and Xhosa. Despite limitations related to translation accuracy, corpus size, and technical constraints, the study validates the feasibility and effectiveness of multilingual sentiment analysis frameworks for low-resource African languages. Future directions emphasize larger and more diverse corpora, explainable AI techniques, and scalable lexicon expansion, providing a roadmap for improving coverage, accuracy, and interpretability in multilingual sentiment modeling. This study establishes a robust, and reproducible methodology that can serve as a foundation for further research and applications in low-resource multilingual NLP.

**Lexicon and Code Availability.**

https://www.kaggle.com/code/stmakhoba/llm-ensemble-on-a-multilingual-lexicon

**Word Frequency Analysis - South African Languages**

*Total Words by Language*

*Unique Words by Language*

*Vocabulary Richness (Unique/Total)*

*Top Words - CILUBA*